\documentclass[runningheads]{llncs}

\usepackage{eccv}

\usepackage{eccvabbrv}

\usepackage{graphicx}
\usepackage{booktabs}
\usepackage{bbm}
\usepackage{xfrac}
\usepackage{wrapfig}
\usepackage{algorithm}
\usepackage{algpseudocode}
\usepackage{algorithmicx}
\usepackage{tcolorbox}
\usepackage{listings}
\definecolor{codegreen}{rgb}{0,0.6,0}
\definecolor{codegray}{rgb}{0.5,0.5,0.5}
\definecolor{codepink}{RGB}{252, 142, 172}
\definecolor{codepurple}{rgb}{0.58,0,0.82}
\definecolor{backcolour}{RGB}{245,245,245}
\lstdefinestyle{mystyle}{
    backgroundcolor=\color{backcolour},   
    commentstyle=\color{magenta},
    keywordstyle=\color{blue},
    numberstyle=\tiny\color{codegray},
    stringstyle=\color{codepurple},
    basicstyle=\fontfamily{\ttdefault}\footnotesize,
    breakatwhitespace=false,         
    breaklines=true,                 
    keepspaces=true,    
    frame=single,
    numbersep=5pt,                  
    showspaces=false,                
    showstringspaces=false,
    showtabs=false,                  
    tabsize=2,
    classoffset=1, 
    keywordstyle=\color{violet},
    classoffset=0,
}
\lstdefinelanguage{JavaScript}{
  keywords={typeof, new, true, false, catch, function, return, null, catch, switch, var, if, in, while, do, else, case, break},
  keywordstyle=\color{blue}\bfseries,
  ndkeywords={class, export, boolean, throw, implements, import, this},
  ndkeywordstyle=\color{darkgray}\bfseries,
  identifierstyle=\color{black},
  sensitive=false,
  comment=[l]{//},
  morecomment=[s]{/*}{*/},
  commentstyle=\color{purple}\ttfamily,
  stringstyle=\color{red}\ttfamily,
  morestring=[b]',
  morestring=[b]"
}
\tcbuselibrary{skins, breakable}
\usepackage[accsupp]{axessibility}  

\usepackage{hyperref}

\definecolor{redhl}{HTML}{FF0000}
\definecolor{greenhl}{HTML}{00FF00}
\definecolor{bluehl}{HTML}{00FFFF}
\definecolor{greyhl}{HTML}{C0C0C0}

\usepackage{orcidlink}

\begin{document}

\title{\underline{S}ee and \underline{T}hink: \underline{E}mbodied Agent in \underline{V}irtual \underline{E}nvironment} 

\titlerunning{STEVE}

\author{
Zhonghan Zhao\inst{1}$^{, \spadesuit}$ \orcidlink{0009-0001-6537-376X} \and
Wenhao Chai\inst{2}$^{, \spadesuit}$ \orcidlink{0000-0003-2611-0008} \and
Xuan Wang\inst{1}$^{, \spadesuit}$ \orcidlink{0009-0007-0073-6893} \and
Boyi Li\inst{1} \orcidlink{0009-0004-7271-3429} \and
Shengyu Hao\inst{1} \orcidlink{0000-0002-8652-8556} \and
Shidong Cao\inst{1} \orcidlink{0009-0005-7132-8753} \and
Tian Ye\inst{3} \orcidlink{0000-0002-8255-2997} \and
Gaoang Wang\inst{1}$^{, \dagger}$ \orcidlink{0000-0002-8403-1538}\\
}
\authorrunning{Z. Zhao et al.}
\institute{Zhejiang University \and University of Washington \and Hong Kong University of Science and Technology (GZ)
}
\maketitle

\begin{abstract}

Large language models (LLMs) have achieved impressive pro-gress on several open-world tasks. Recently, using LLMs to build embodied agents has been a hotspot. 
This paper proposes STEVE, a comprehensive and visionary embodied agent in the Minecraft virtual environment. STEVE comprises three key components: vision perception, language instruction, and code action. Vision perception involves interpreting visual information in the environment, which is then integrated into the LLMs component with agent state and task instruction. Language instruction is responsible for iterative reasoning and decomposing complex tasks into manageable guidelines. Code action generates executable skill actions based on retrieval in skill database, enabling the agent to interact effectively within the Minecraft environment. We also collect STEVE-21K dataset, which includes 600$+$ vision-environment pairs, 20K knowledge question-answering pairs, and 200$+$ skill-code pairs. We conduct continuous block search, knowledge question and answering, and tech tree mastery to evaluate the performance.
Extensive experiments show that STEVE achieves at most $1.5 \times$ faster unlocking key tech trees and $2.5 \times$ quicker in block search tasks.

\end{abstract}

\newcommand\blfootnote[1]{ 
\begingroup 
\renewcommand\thefootnote{}\footnote{#1}
\addtocounter{footnote}{-1}
\endgroup 
}
{
\blfootnote{
 {{$^\spadesuit$}} Equal contribution: \href{mailto:zhaozhonghan@zju.edu.cn}{\color{black}{zhaozhonghan@zju.edu.cn}}, \href{mailto:wchai@uw.edu}{\color{black}{wchai@uw.edu}}, \href{mailto:xuanw@zju.edu.cn}{\color{black}{xuanw@zju.edu.cn}}.
 
 $^\dagger$ Corresponding author: \href{mailto:gaoangwang@intl.zju.edu.cn}{\color{black}{gaoangwang@intl.zju.edu.cn}}.
}
}

\section{Introduction}
Designing agents that demonstrate intelligent behavior and adaptability in open-world settings has been a longstanding and significant challenge in the field of artificial intelligence~\cite{kolve2017ai2,savva2019habitat,zhao2023survey,deng2023citygen,deng2024citycraft}. However, recent progress in the development of large language models (LLMs)~\cite{chatgpt,touvron2023llama} has exhibited their potential as versatile, general-purpose assistants. Recent innovations in agent design~\cite{yuan2023plan4mc,wang2023voyager,wang2023describe,zhu2023ghost} have effectively harnessed these advanced LLMs, tapping into their extensive world knowledge and reasoning abilities. This development has paved the way for agents, that are autonomously driven, to formulate and implement strategies and actions across a diverse array of skills and tasks in diverse open-world environments.

\begin{figure}[t]
\centering
\includegraphics[width=0.95\textwidth]{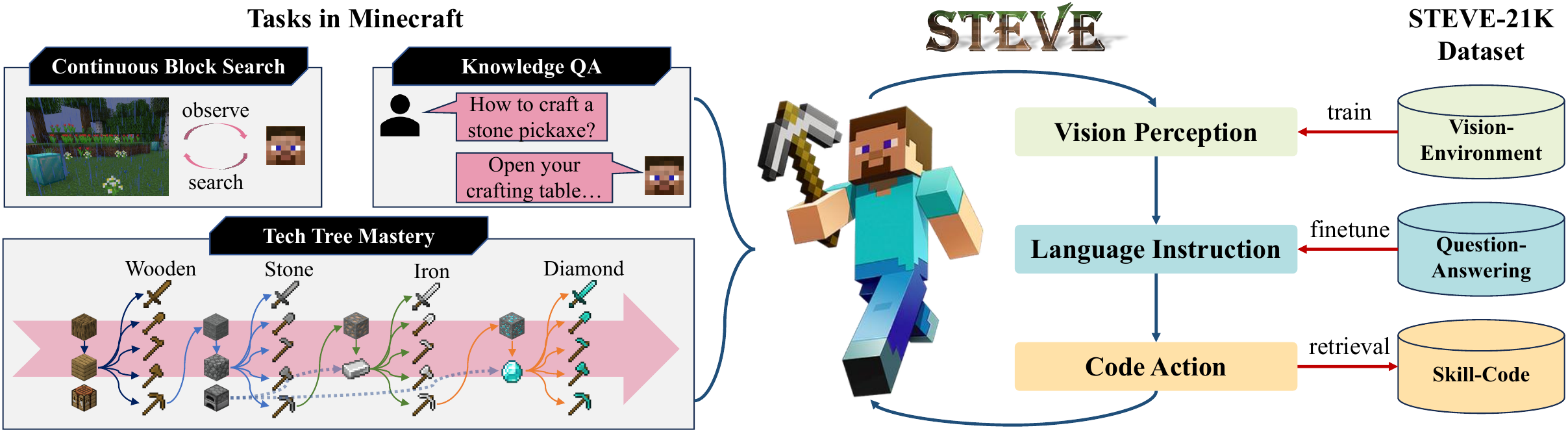}
\captionof{figure}{STEVE integrates the three parts: Vision Perception, Language Instruction, and Code Action, supported closely by our proposed STEVE-21K. It demonstrates commendable performance on Continuous Block Search, Knowledge QA, and Tech Tree Mastery.}
\label{fig:teaser}
\end{figure}

In many open-world settings, like Minecraft, contemporary agents predominantly use LLMs for their textual interactions. However, this reliance on text for communication poses considerable limitations in their interactions within these worlds including low-level regrading cases~\cite{vasluianu2024ntire,jiang2023five,ye2023sequential,ye2022perceiving}. Minecraft, with its expansive and interactive sandbox environment~\cite{guss2019minerl,fan2022minedojo}, demands a variety of skills from agents, ranging from crafting basic items to executing complex tasks. Yet, agents driven by LLMs often generate unpredictable outputs. The effectiveness of their interactions is largely contingent on meticulously crafted prompts~\cite{huang2022inner}, designed to align the LLM’s understanding with the environmental context and the intended objectives. This process of prompt engineering is not only laborious but also fails to meet the goal of fostering autonomous, self-directed agents. Furthermore, textual communication has its limitations in naturally conveying certain concepts of the world, like crafting recipes, which are often more effectively communicated through vision.

Players have the distinct capability to assimilate and convey information using both visual and textual channels, significantly enhancing our interactions with the world around us. Yet, the integration of LLM-based agents with multimodal inputs in open-ended environments remains an under-explored area. 
STEVE in STEVE-Series~\cite{zhao2024steve, zhao2024hierarchical, zhao2024we}, is named after the protagonist of the game ``Minecraft,''. It is our proposed framework to build an embodied agent based on the vision model and LLMs within an open world, as illustrated in~\cref{fig:teaser}. STEVE harnesses a vision model to perceive its surroundings visually, coupled with an LLM to strategize and plan actions. This model represents a leap forward in agent design, combining these two input modes, vision, and text, to offer a more nuanced and comprehensive understanding of the environment and practical and executable skills.

Our key contributions are outlined as follows:
\vspace{-6pt}
\begin{itemize}
    \item [$\bullet$] We propose STEVE, an embodied agent in virtual environment, consists of vision perception, language instruction, and code action, achieving $1.5 \times$ faster unlocking of key tech trees and is $2.3 \times$ quicker in block search tasks compared to previous state-of-the-art methods.
    \item [$\bullet$] We present STEVE-7B/13B, a series of large language model obtained by fine-tuning with Minecraft knowledge question-answering pairs from Llama-2-7B/13B.
    \item [$\bullet$] We collect STEVE-21K dataset, including 600+ vision-environment pairs, 20K knowledge question-answering pairs, and 200+ skill-code pairs, for justifying the effective performance of STEVE.
\end{itemize}
\section{Related Works}

\subsection{Intelligent Agent in Minecraft}
As an open-ended sandbox game, Minecraft has always been an ideal setting for testing the performance of intelligent agents~\cite{johnson2016malmo,hofmann2019minecraft}. The agents are required to autonomously perform various tasks in Minecraft, such as chopping trees, crafting tools, and mining diamonds. At the beginning, much of the works focus on exploring reinforcement learning~\cite{lin2021juewu,mao2022seihai,skrynnik2021hierarchical,lifshitz2023steve} or imitation learning~\cite{amiranashvili2020scaling,baker2022video}, without satisfactory performance. VPT~\cite{baker2022video} and MineDojo~\cite{fan2022minedojo} collect internet-scale datasets for their model pre-training. More specifically, VPT offers the exciting possibility of directly learning to act during video pre-training and using these learned behavioral priors as extremely effective exploration priors for reinforcement learning.
Yet, recent works found that the pre-trained LLMs could serve as a strong ``mind'' that provides planning ability to the agents. Voyager~\cite{wang2023voyager} leverages GPT-4~\cite{openai2023gpt4} as both a high-level planner and a low-level action code generator. Plan4MC~\cite{yuan2023plan4mc} proposes a skill graph pre-generated by the LLMs. DEPS~\cite{wang2023describe}, an interactive planning method based on LLMs,  addresses multi-step reasoning issue in open-world planning. GITM~\cite{zhu2023ghost} develops a set of structured actions and leverages LLMs to generate action plans for the agents to execute,  achieving impressive results in various tasks. 

\subsection{Embodied Multimodal Model}

Embodied agent operates within various environment by synthesizing sensory perceptions and physical actions supported by computational intelligence. This synthesis enables the agent to undertake a variety of tasks, achieving specific objectives. Its key areas of application are diverse, including Navigation~\cite{wijmans2019dd,yu2021sound,du2020vtnet,kwon2023renderable,chen2021history,moudgil2021soat}, Embodied Question Answering~\cite{das2018embodied,yu2019multi,datta2022episodic}, Active Visual Tracking~\cite{luo2018end,zhong2021towards,luo2019end,zhong2019ad}, and Visual Exploration~\cite{liu2022symmetry,dean2020see,chen2018learning}. The field is evolving rapidly with the development of Large Language Models (LLMs)~\cite{song2022llm} and Multimodal LLMs (MLLMs)~\cite{alayrac2022flamingo,zhu2023minigpt,li2023otter,li2022blip,li2023blip,gong2023multimodal,lyu2023macaw,ye2023mplug,dai2023instructblip,wang2023visionllm,liu2023visual,maaz2023video,su2023pandagpt,gao2023llama}, integrating multiple modalities for more effective processing. A prime example of this innovation is PaLM-E~\cite{driess2023palm}, a sophisticated multimodal model with 562B parameters, adept at a broad spectrum of embodied tasks and demonstrating exceptional capabilities in visual reasoning.

\subsection{Large Language Model with Equipped Tools}

While Large Language Models (LLMs) demonstrate impressive skill in tackling novel tasks via prompt-based instructions, they often face challenges in areas where simpler models or tools excel, like mathematical calculations or identifying palindromes. However, LLMs' potential is significantly expanded when integrated with other modality-specific models, such as those for vision or audio, enabling multi-modal capabilities~\cite{chai2022deep,lu2023chameleon}. Innovations like Toolformer~\cite{schick2023toolformer} demonstrate LLMs' self-learning to utilize tools through finetuning with extensive API call samples. Visual ChatGPT~\cite{wu2023visual} extends this by integrating various visual foundation models, facilitating interactive user experiences with ChatGPT. Similarly, HuggingGPT~\cite{shen2023hugginggpt} presents a framework that harnesses LLMs to link diverse models from Hugging Face for task resolution. AutoGPT~\cite{autogpt} is an open-source application that broadens GPT-4's capabilities with internet access, memory management, and plug-ins. The recent introduction of MovieChat~\cite{song2023moviechat, song2024moviechat+} brings a memory mechanism to MLLM, enhancing its performance in video understanding tasks. Furthermore, LLMs can be used for goal planning, analogous to language translation~\cite{xie2023translating}. This evolving landscape suggests that tool-equipped LLMs could forge a new paradigm in AI solution design.
\section{Method: STEVE}

\begin{figure*}[t]
    \centering
    \includegraphics[width=1\linewidth]{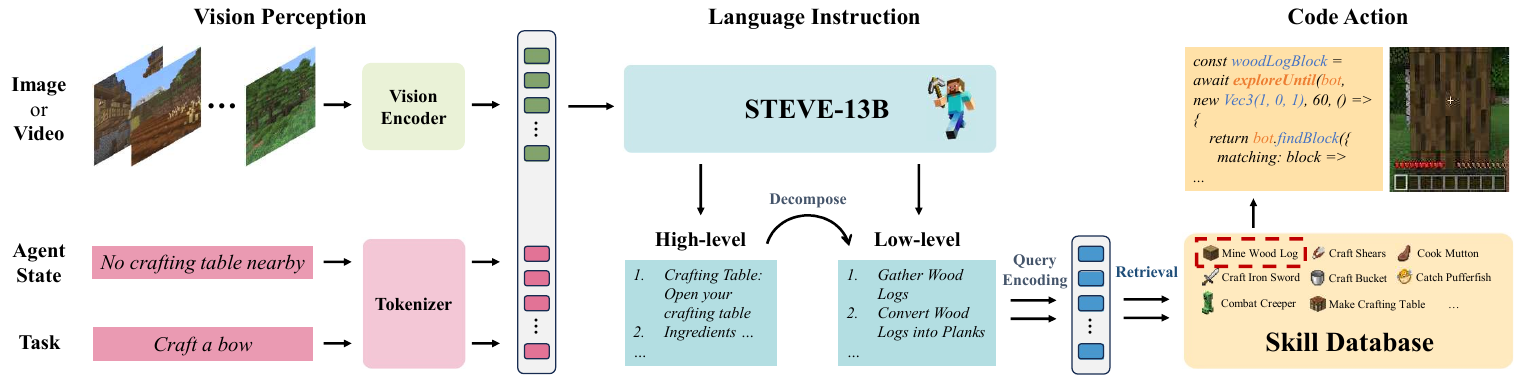}
    \caption{\textbf{STEVE framework.} The Vision Perception part takes images or videos, encodes them into tokens, and combines them with Agent State and Task tokens as input. The STEVE-13B in the Language Instruction part is used for automatic reasoning and task decomposition, and it calls the Skill Database in the form of the Query to output code as action.}
    \label{fig:framework}
\end{figure*}

\subsection{Overview}

As shown in~\Cref{fig:framework}, STEVE $\mathcal{F}$ is an LLM-based multi-modal autonomous system for embodied agents in Minecraft, which can use visual status $X^v$ in the form of images or videos and agent state $X^s$ to manage and execute complex task $X^t$ for executive code action $\mathbf{a}^c$:

\begin{equation}
\mathbf{a}^c = \mathcal{F}(X^v, X^s, X^t) = \mathcal{A}^c(\mathcal{I}^l(\mathcal{P}^v(X^v, X^s, X^t)))
\end{equation}
where STEVE $\mathcal{F}$ integrates Visual Perception $\mathcal{P}^v$ into the Language Instruction $\mathcal{I}^l$ with large language model (LLM) and combines them with Code Action $\mathcal{A}^c$, a skill retrieval method to execute actions, details are demonstrated as follows.

\begin{itemize}
    \item [$\bullet$] \textbf{Vision Perception $\mathcal{P}^v$~(Section~\ref{sec:Perception})}; a vision encoder to interpret visual information of the environment, such as blocks and entities and a text tokenizer for agent state $s$ and task $t$.
    \item [$\bullet$] \textbf{Language Instruction $\mathcal{I}^l$~(Section~\ref{sec:Instruction})}: a powerful language model system fine-tuned specifically for Minecraft content using LLaMA2-13B~\cite{gao2023llama}. This model enables adaptive interaction for iterative reasoning and step-by-step decomposition.
    \item [$\bullet$] \textbf{Code Action $\mathcal{A}^c$~(Section~\ref{sec:Action})}: a skill retrieval method based on our skill-code database.
\end{itemize}

\subsection{Vision Perception $\mathcal{P}^v$}\label{sec:Perception}
The vision perception part includes a vision encoder $E^v$ and a text tokenizer $T$, which converts the visual status $X^v$, agent state $X^s$, and task $X^t$ to the text-space tokenizer representation $\mathbf{Y} = \{Y^v, Y^s, Y^t\}$:
 
\begin{equation}
\mathbf{Y}_i = \mathcal{P}^v(X^v_i, X^s_i, X^t_i), \mathcal{P}^v = \{E^v, T\},
\end{equation}
where the vision encoder $E$, the visual branch of EfficientFormer~\cite{li2022efficientformer}, encodes visual status $X^v_i$ at each step $i$ into visual tokens $Y^v=\{y^v_1, y^v_2,..., y^v_n\} \in \mathbbm{R}^{n \times d}$, where $n$ denotes the number of visual tokens $y^v$ and $d$ is the dimensionality of each token. The text tokenizer converts agent state $X^s_i$ and task $X^t_i$ at each step $i$ in the form of text into textual tokens $Y^s$ and $Y^t$. Note that visual tokens are amalgamated with textual tokens representing the agent's current state (\eg, health, inventory, \etc) and the task description. This is accomplished using a tokenizer that maps state variables and task parameters to a tokenized form. The resultant unified token set serves as a comprehensive representation of the current situational context.

\subsection{Language Instruction $\mathcal{I}^l$}\label{sec:Instruction}
The Language Instruction $\mathcal{I}^l = \{\mathcal{P}_l, \mathcal{C}_r, \mathcal{C}_u, \mathcal{D}_s\}$, which consists of Planner $\mathcal{P}_l$, Critic $\mathcal{C}_r$, Curriculum $\mathcal{C}_u$, and Describer $\mathcal{D}_s$, \textit{i.e.}, four independent LLM-based agents with different functions. They formulate high-level task guidelines, refine strategies through feedback, facilitate continuous learning and adaptation through a curriculum of complex tasks, and finally decompose strategic guidelines into the executable low-level textual action step $\mathbf{a^s}$, $\{\mathbf{a^s} \sim\mathcal{I}^l(\cdot|\mathbf{Y}_i)\}_{i=0}^{N}$ towards efficient completion of the task.

Note that the Planner $\mathcal{P}_l$, Critic $\mathcal{C}_r$, Curriculum $\mathcal{C}_u$ and Describer $\mathcal{D}_s$ are based on STEVE-7B/13B, a powerful language model derived from LLaMA-2-13B~\cite{gao2023llama}, fine-tuned specifically on Minecraft-related content from the STEVE-21K. This model's expertise covers a broad spectrum of game-specific knowledge areas on worlds, entities, player mechanics, survival, and even game practical experience. Finally, they have the performance of their different functions:

\begin{itemize}
    \item [$\bullet$] \textbf{Planner $\mathcal{P}_l$:} formulating comprehensive guidelines and executive plans that align with the overarching objectives of the task.
    \item [$\bullet$] \textbf{Critic $\mathcal{C}_r$:} evaluating the planner decisions, providing feedback that can refine strategies.
    \item [$\bullet$] \textbf{Curriculum $\mathcal{C}_u$:} facilitating continuous learning and adaptation for action agents by engaging with a series of progressively complex tasks.
    \item [$\bullet$] \textbf{Describer $\mathcal{D}_s$:} distilling the extensive data into a concise summary, making it more manageable and interpretable.
\end{itemize}

\paragraph{Iterative reasoning.}
The STEVE-13B receives a stream of tokens that encode the current visual scene, the agent's state, and the task's textual description. STEVE-13B interprets this rich context to undertake complex reasoning. The model initiates the reasoning process by constructing a series of high-level strategies that outline the pathway to task completion.
Considering all gameplay elements, the reasoning mechanism is akin to an experienced player who can visualize the end game and chart a course to victory. This approach ensures the plans are reactive and proactive, allowing the agent to anticipate and mitigate future challenges. However, most strategies are high-level and abstract therefore, they often require step-by-step decomposition to derive executable guidelines.

\paragraph{Decomposition.}
The decomposition process makes the complex strategies break down into simple, low-level guidelines that can be directly mapped to actions in Minecraft. It is similar to how a high-level command like ``build a shelter'' is divided into actionable instructions like ``collect wood'', ``craft planks'', and ``place blocks''. The granular steps are structured to provide explicit instructions that the game engine can readily interpret. This systematic breakdown from high-level reasoning to low-level actionable steps is the hallmark of the STEVE system, enabling the embodied agent to interact with the Minecraft environment in a meaningful and goal-oriented manner. Through this intricate process of reasoning and decomposition, STEVE embodies the cognitive capabilities required for sophisticated task management in virtual environments.


\paragraph{Curriculum learning with memory.} 
We draw inspiration from the lifelong learning strategy utilized in many reinforcement learning problems~\cite{wang2023voyager} in both closed-world and open-world settings. We start by creating a set of tasks that serve as a curriculum for agents to explore the environment. During this process, STEVE generates plans, interacts with the surroundings, learns from mistakes, and stores all these experiences in memory. Next, we evaluate STEVE on various tasks after this learning stage. Consequently, STEVE can produce better plans with its memory teaming up with the planning experiences. We use this as the default setting for all tasks in our experiments.

\paragraph{Continous learning with summarization.}~\label{para:COS}
We've observed that the learning process, where the memory is being filled, can continue throughout the gameplay. The agent can gradually acquire more skills as the gameplay progresses and more experiences are gained. However, as the memory gets larger, it becomes difficult to understand the game's situation and interact with the game slowly. To tackle these challenges, we have implemented the Chain of Summarization method~\cite{ma2023large}. By finding better references, we can improve our ability to handle challenging tasks, such as ``Diamond Tool'' in the tech tree, including obtaining materials and creating diamond tools. This will lead to a higher success rate, as shown in Section~\ref{subsec:eval_results}. Additionally, curriculum learning with memory allows for in-context lifelong learning without needing gradient updates.

\subsection{Code Action $\mathcal{A}^c$}\label{sec:Action}

The code action part $\mathcal{A}^c$ is the execution phase, where the STEVE system converts planned, decomposed guidelines into concrete actions within the Minecraft environment. This process leverages a specialized skill database that pairs code snippets with their descriptions and relevant metadata, encoded as vectors $\mathbf{v^s}$ for efficient retrieval. The transition from language instruction to executable code is achieved through the retrieval process $\mathcal{R}$:

\begin{equation}
    \mathcal{R}(\mathbf{q}, \mathbf{v}) = \sigma(\mathbf{q}, \mathbf{v}), \mathbf{q} = E^q(\mathbf{a^s}),
\end{equation}
where $E^q$ and $\sigma$ are query encoding and cosine similarity matching. Each low-level textual action step $\mathbf{a^s}$ derived from the Language Instruction is encoded into a query $\mathbf{q}$. This encoding captures the essence of the action to be performed in a format that the skill database can understand. Once the queries $\mathbf{q}$ are encoded, the system computes similarities between the query vectors and the code snippets vectors $\mathbf{v}$ stored in the database to determine which skill best matches the required action.

\subsection{Training}

To reduce the training overhead to a certain extent, we adopt a two-stage training method, the first stage being a warm-up on the question-answering pairs of STEVE-21K to ensure a certain degree of instruction capability. The second stage is simulated in the environment, and the Expert LLM $\mathcal{E}_p$ integrated with GPT-4~\cite{openai2023gpt4} generates instructions for the same situation to modify our model.

In training, we use either a single-round conversation $\{X^Q, X^A\}$ or a multi-round conversation $\{X^Q_i, X^A_i\}_{i\leq N}$, where $N$ is the total number of rounds in the conversation. We use the negative log-likelihood objective over the prediction tokens with training and finetuning:
\begin{equation}
    \mathcal{L}(\theta)=-\sum_{j=1}^{L} \log \mathcal{F}_{\theta}(Y_j|X^v, \hat{Y}_{1:j-1}),
\end{equation}
where $Y$ and $\hat{Y}$ refer to the non-vision input and target token sequences, $\theta$ represents the model parameters, and $L$ represents the length of the target sequence. The vision input $X^v$ varies depending on the input step. We only consider the answer tokens $X_A$ when computing the loss to ensure that the model focuses on generating coherent responses.


\noindent\textbf{Offline warm-up.}
In the first stage, we finetune the STEVE-7B/13B only on the single-round question-answering pairs of STEVE-21K to ensure a certain degree of instruction capability, as shown in~\cref{table:exp_inst}.

\noindent\textbf{Online finetuning.}
In the second stage, we simulate the warmed-up STEVE-7B/13B in Minecraft, and both train the vision encoder $E^v$ and finetune the STEVE-7B/13B simultaneously. 

It is necessary to obtain the environment information to train the vision encoder, denoted as $l = R_T(X^v)$, where $l$ is the label of seen blocks and entities, $R_T$ is the Ray Tracing method~\cite{whitted2005improved} to eliminate all content outside the agent's perspective screen. We train the vision encoder $E^v$ once we have obtained the environment information. To finetune the warmed-up STEVE-7B/13B in simulation, we give both the STEVE-7B/13B and the Expert LLM the same input, and we consider the output of the Expert LLM as the ground truth label.

After 5,000 simulations, we collect all the context information from successful runs, including sequences of vision and question-answering chat flow as the Vision-Environment part of STEVE-21K, as mentioned in Section~\ref{sec:dataset}.
\begin{figure*}[t]
    \centering
    \includegraphics[width=1\linewidth]{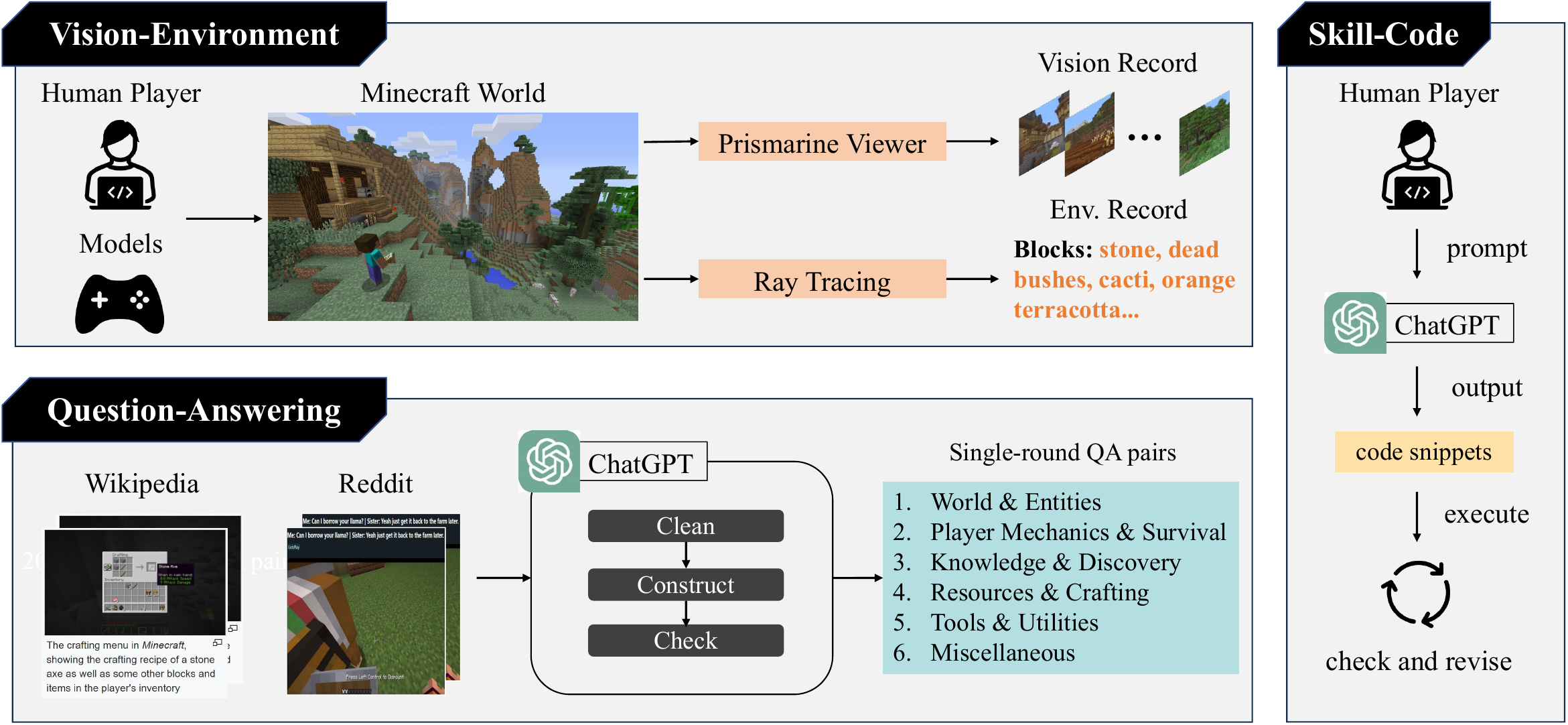}
    \caption{\textbf{STEVE-21K collection pipeline.} In the Vision-Environment section, STEVE-13B plays the game according to specified tasks defined by the human player, collecting visual information through prismarine-viewer and capturing environmental information from the screen using Ray Tracing~\cite{whitted2005improved}. Note that the language instruction task is also performed during the collection phase. We simultaneously record and save the chat flow from the reasoning and decomposition stages. In the Question-Answering section, we obtain information from the Minecraft-Wiki and Reddit forums and use GPT-3.5 to clean the data into single-round QA pairs. In the Skill-Code section, we use GPT-3.5 combined with the human player's code to synthesize code snippets and then check and revise them in the game environment.}
    \label{fig:dataset_pipeline}
\end{figure*}
\begin{figure}[t]
    \centering
    \includegraphics[width=0.95\linewidth]{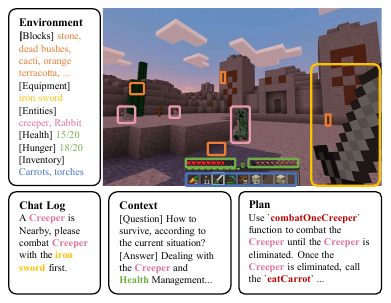}
    \caption{\textbf{Example of Vision-Environment pairs.} It represents the data format of the Vision-Environment pairs in our STEVE-21K dataset: including visual signals, environmental information, Chat Log, Context QA pairs, and planning in actual tasks.}
    \label{fig:dataset_illustration}
\end{figure}

\section{Dataset: STEVE-21K}\label{sec:dataset}
As shown in~\cref{fig:dataset_pipeline}, STEVE-21K has been meticulously compiled, featuring a diverse collection of Vision-Environment pairs, Question-Answering pairs, and a Skill-Code database. 

\noindent\textbf{Vision-Environment} pairs contain 600 pairs of first-person perspective videos from Minecraft gameplay across six different terrains (including forest, desert, coastal, \etc), along with corresponding environmental block entities in the field of vision and context information for each timestamp. Additionally, all pairs are oriented around executing the skill-related task supported by Skill-Code part mentioned in Section \ref{sec:skill_database}. We employ the STEVE-7B/13B model to enable robots to plan and execute actions autonomously based on tasks defined by human supervisors. We record the video of the agent operation, the environment information, and all the corresponding chatflow. 

\noindent\textbf{Question-Answering} pairs contain 20K question-answering pairs from the Minecraft-Wiki and Reddit corpus across six data types partly sourced from~\cite{fan2022minedojo}. 
The pairs are organized into instruction, input, and output triplets and used to train STEVE-13B. The GPT-3.5~\cite{chatgpt} is employed to derive meaningful single-round question-answer pairs, and LoRA~\cite{hu2021lora} is incorporated during the fine-tuning process for efficient resource allocation.

\noindent\textbf{Skill-Code}~\label{sec:skill_database} pairs contain 210 skill execution scripts with descriptions, covering 8 skill types including collecting, crafting, exploration~\etc. The code part is collected by manual coding. We use GPT-3.5~\cite{chatgpt} to describe all codes and utilize langchain vectordb to give all pairs a database vector. 

\section{Experiments}
\subsection{Experimental Setup} 
We train STEVE-7B/13B, which is finetuned from \text{LLaMA-2}~\cite{gao2023llama} with the Queston-Answering pair in STEVE-21K dataset for warm-up and simulation context data from successful runs. We use LoRA~\cite{hu2021lora} for finetuning process. Note that the process is to adjust STEVE-13B to work on correct simulation knowledge while remaining adapted to visual perception.
In the text part, we set all temperatures to 0 except for the task proposal, which uses 0.9 to encourage task diversity. The vision unit is based on \text{EfficientFormerV2-S0}~\cite{li2022efficientformer}, which is trained on the Vision-Environment part of our STEVE-21K dataset. Our simulation environment is built on top of MineDojo~\cite{fan2022minedojo} and leverages Mineflayer~\cite{mineflayer}. We use GPT-4-0613 for all GPT-4 models used in Voyager~\cite{wang2023voyager} and code generation tasks.

\subsection{Baselines}
As no vision-based LLM-driven agents are immediately operable in Minecraft, we selected several algorithms as baselines that extract information from a system's backend, differing significantly from real-world applications.

\paragraph{AutoGPT}\cite{autogpt} is an NLP automation tool that decomposes a high-level goal into executable subgoals in MineDojo, aligning with our experimental framework. Our setup, AutoGPT, powered by GPT-4~\cite{openai2023gpt4}, processes agent states, environment feedback, and execution errors to manage subgoal execution.

\paragraph{Voyager}\cite{wang2023voyager} relies solely on textual grounding for perception and features a long-term procedural memory with a hierarchical library of code-based procedures, allowing the integration of simple skills into complex behaviors. Proficient in environment exploration and tech tree mastery, Voyager uses GPT-4~\cite{openai2023gpt4} for processing background text in embodied agents, with a lesser emphasis on visual perception.

\subsection{Evaluation Results}~\label{subsec:eval_results}
\paragraph{Continuous block search.} 
\begin{wraptable}{r}{0.5\textwidth}
\centering
\caption{\textbf{Comparison on continues block search task.} \# Iters stand for average iterations to find 10 diamonds (max 100). \# Blocks stand for average diamonds found in 100 iterations.
}
\begin{tabular}{l|cc}
\toprule
Method & \# Iters~($\downarrow$) & \# Blocks~($\uparrow$) \\ 
\midrule
AutoGPT~\cite{autogpt} & N/A & 7\\
Voyager~\cite{wang2023voyager} & 35 & 26\\
\textbf{STEVE} & \textbf{14} & \textbf{67}\\
\bottomrule
\end{tabular}
\label{table:exp_visual}
\end{wraptable}
As shown in~\cref{table:exp_visual}, we experiment with block-searching tasks to assess the agent's exploratory capabilities and proficiency in locating specified blocks. Diamond blocks are placed at every 16-block interval across the land map. The agent's objective is to identify as many blocks as possible within the fewest iterations, which indicates the method's efficiency. 
As shown in~\cref{fig:vision_show}, enriching information through visual perception significantly enhances the efficiency of search and exploration tasks, leading to more effective world exploration.

\begin{table}[t!]
\centering
\caption{
Comparison on \textbf{tech tree mastery} task. The values presented are in fractions, representing successful trials out of three attempts. A score of 0/3 signifies the method's inability to progress within the tech tree after a maximum of 160 prompting iterations. The reported numbers denote the average iterations across three trials. Lower iteration values indicate higher efficiency of the respective method.}
\resizebox{\linewidth}{!}{
\begin{tabular}[c]{l|cccc}
\toprule
Method & Wooden Tool & Stone Tool  & Iron Tool  & Diamond Tool \\ 
\midrule
AutoGPT~\cite{autogpt} & $92 \pm 72$ $(\sfrac{3}{3})$ & $94 \pm 72$ $(\sfrac{3}{3})$ & $135 \pm 103$ $(\sfrac{3}{3})$ & N/A $(\sfrac{0}{3})$ \\
Voyager~\cite{wang2023voyager} & $6\pm 2$ $(\sfrac{3}{3})$ & $11 \pm 2$ $(\sfrac{3}{3})$ & $21\pm 7$ $(\sfrac{3}{3})$ & \textbf{102} $(\sfrac{1}{3})$ \\
\textbf{STEVE} & \textbf{4} $\pm$ \textbf{1} $(\textbf{\sfrac{3}{3}})$ & \textbf{8} $\pm$ \textbf{1} $(\textbf{\sfrac{3}{3}})$ & \textbf{15} $\pm$ \textbf{2} $(\textbf{\sfrac{3}{3}})$ & 106 $\pm$ 12 $(\textbf{\sfrac{3}{3}})$ \\
\bottomrule
\end{tabular}
}
\label{table:tech_tree}
\end{table}
\definecolor{ST}{RGB}{0,112,192}
\definecolor{VY}{RGB}{255,192,0}
\definecolor{AG}{RGB}{192,0,0}
\begin{figure}[t]
    \centering
    \includegraphics[width=0.8\linewidth]{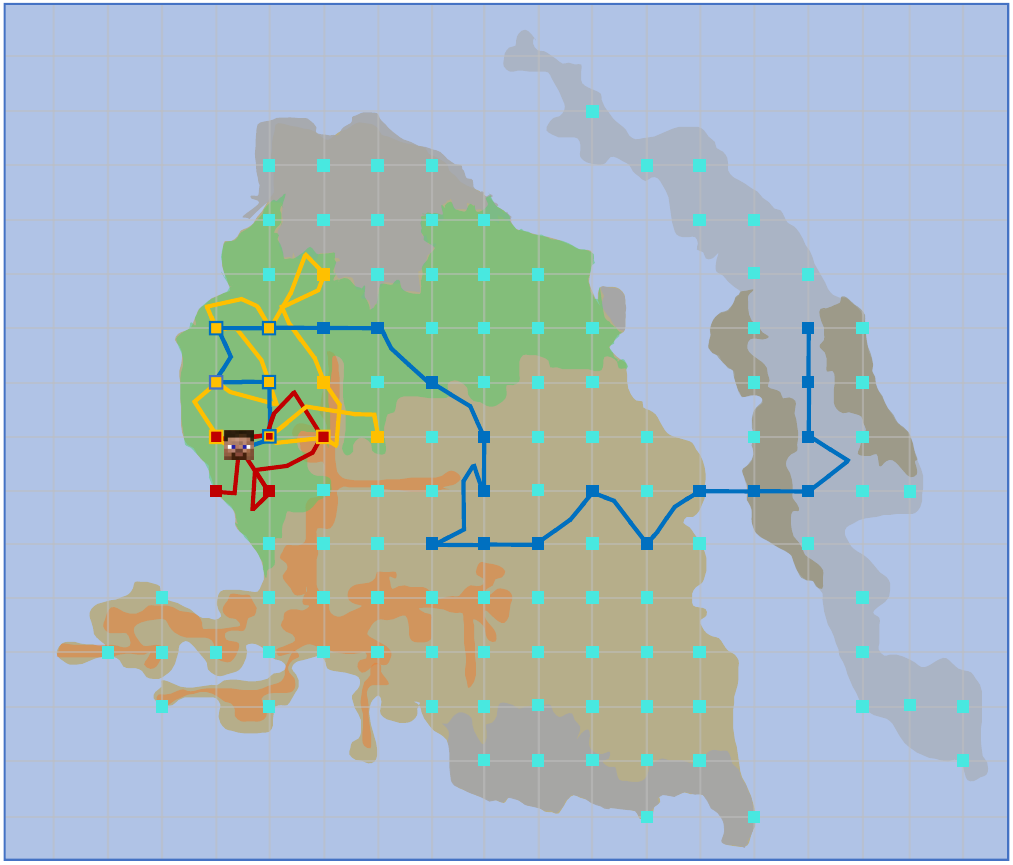}
    \caption*{
    \parbox[c][8pt][l]{8pt}{\colorbox{ST}{}}STEVE
    \quad
    \parbox[c][8pt][l]{8pt}{\colorbox{VY}{}}Voyager
    \quad
    \parbox[c][8pt][l]{8pt}{\colorbox{AG}{}}AutoGPT
    }
    \caption{\textbf{Schematic of the continuous block search task.} We capture an asynchronous segment with each method 30 iterations from the experiments for illustration. The reason we choose diamond blocks is that they are not naturally occurring in the given context, making them easily distinguishable from other blocks.}
    \label{fig:vision_show}
\end{figure}

\paragraph{Knowledge question and answering.} 
\begin{table*}[t!]
\centering
\caption{\textbf{Quantitive comparison on knowledge question and answering task}. Questions, model-generated responses, and ground truth inputs are evaluated in GPT-4~\cite{chatgpt}, Claude-2~\cite{claude} and human blind rating rated on a scale of 0 to 10; The scores above are the average of them. Higher scores indicate greater alignment of the generated answers with the ground truth. Wld., Ent., Mech., Surv., Know., Disc., Res., Craft., Tools, Util., Miscell. stand for World, Entities, Player Mechanics, Survival, Knowledge, Discovery, Resources, Crafting, Tools, Utilities and Miscellaneous.}
\resizebox{\linewidth}{!}{
\begin{tabular}[c]{l|cccccc|c}
\toprule
Method  & Wld. \& Ent. & Mech. \& Surv. & Know. \& Disc. & Res. \& Craft. & Tl. \& Util. & Miscell. & Overall\\ 
\midrule
Llama2-7B  & 6.44 & 6.68 & 6.58 & 6.42 & 6.80 & 6.96 & 6.56\\
Llama2-13B & 6.93 & 6.95 & 6.77 & 6.77 & 6.98 & 6.64 & 6.89\\
\midrule
\textbf{STEVE-7B} & 7.99 & 7.88 & 7.84 & 7.95 & 7.93 & 7.82& 7.94\\
\textbf{STEVE-13B} & \textbf{8.14} & \textbf{8.13} & 8.03 & \textbf{8.15} & \textbf{8.12} & 7.72 & \textbf{8.12}\\
\midrule
GPT-4 & 8.06 & 8.07 & \textbf{8.07} & 7.92 & 8.09 & \textbf{8.21} & 8.04\\
\bottomrule
\end{tabular}
}
\label{table:exp_inst}
\end{table*}
\begin{figure*}[t]
    \centering
    \includegraphics[width=\linewidth]{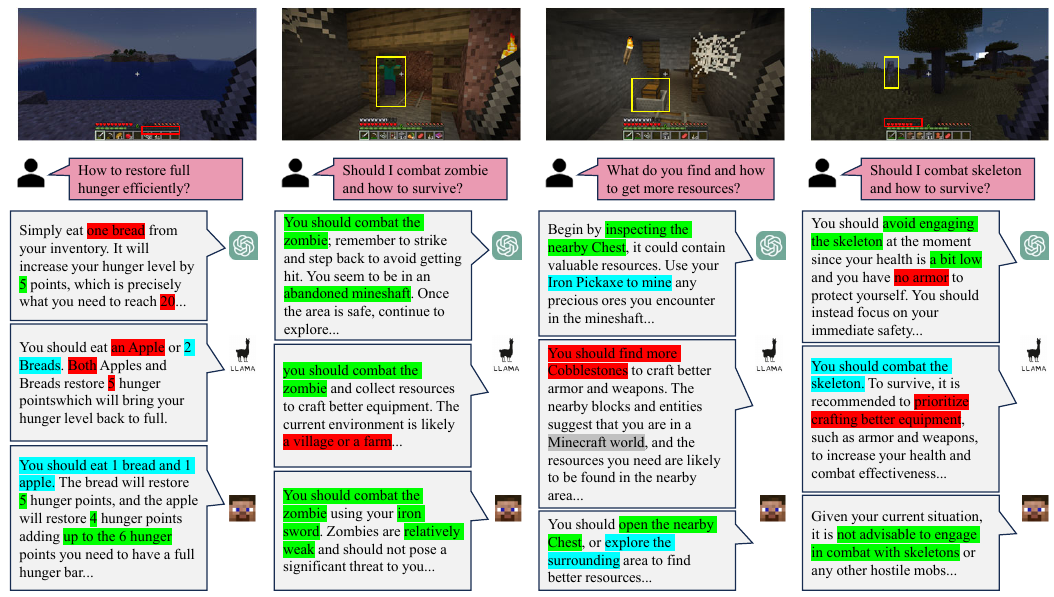}
    \caption{\textbf{Qualitative comparison on knowledge question and answering tasks.} \colorbox{greenhl}{Green}(\colorbox{redhl}{Red}) highlights the correct or good (wrong or bad) answer. \colorbox{bluehl}{Blue} indicates the suboptimal answer. \colorbox{greyhl}{Grey} indicates the meaningless answer.}
    \label{fig:qa}
\end{figure*}

Using a validation dataset, we established a question-answering database to evaluate our model's performance on Minecraft-related queries. Responses from each model are rated blindly by GPT-4, Claude-2, and human participants based on accuracy, relevance, and detail. Responses are initially checked for accuracy, and the comprehensive evaluation yields a score ranging from 0 to 10, with higher scores indicating superior performance.

As detailed in~\cref{table:exp_inst}, we tested the instructional capabilities of various LLM models using the STEVE-21k dataset, divided into $18,622$ training and $1,000$ testing samples. The STEVE-7B and STEVE-13B models surpassed LLaMA2 in all metrics, with STEVE-13B scoring the highest at 8.12, demonstrating its exceptional understanding of Minecraft-related queries. This indicates that STEVE models, optimized specifically for Minecraft content, perform better in knowledge-intensive tasks.

Our results confirm that larger, domain-specifically tuned models like STEVE-13B outperform broader models like GPT-4, underscoring the benefit of specialized fine-tuning for domain-specific applications.

\paragraph{Tech tree mastery.}
As shown in~\cref{table:tech_tree}, we experiment on the Minecraft tech tree mastery to test the agent's ability to craft and use a hierarchy of tools. 
Progressing through this tree (wooden tool $\rightarrow$ stone tool $\rightarrow$ iron tool $\rightarrow$ diamond tool) requires the agent to master systematic and compositional skills. 
As to the wooden, stone, and iron levels of the tech tree, STEVE achieves remarkable efficiency: $23 \times$, $11.8 \times$, and $9 \times$ faster than AutoGPT~\cite{autogpt}, and $1.5 \times$, $1.4 \times$, and $1.3 \times$ faster than Voyager~\cite{wang2023voyager}.
STEVE has achieved the diamond level, as shown in~\cref{table:tech_tree}. Its performance slightly lags behind Voyager~\cite{wang2023voyager}, which also uses GPT4~\cite{openai2023gpt4} for critical inference. However, STEVE is more cost-effective, starting with lower initial performance. It includes a vision unit, prioritizing visual data over background information, offering distinct advantages. Additionally, we observed a decrease in performance when using a basic skill database.

\subsection{Ablation Study}
\begin{table}[t!]
\centering
\caption{
\textbf{Ablation studies for the tech tree mastery.} STEVE~(Ours) is the STEVE-13B version. The 0/3 score means the method can't progress beyond 160 iterations in the tech tree.
}
\resizebox{0.9\linewidth}{!}{
\begin{tabular}[c]{l|cccc}
\toprule
Method & Wooden Tool & Stone Tool  & Iron Tool  & Diamond Tool \\ 
\midrule
w/o vision unit & $11 \pm 5$ $(\sfrac{3}{3})$ & $27 \pm 5$ $(\sfrac{3}{3})$ & $46 \pm 11$ $(\sfrac{3}{3})$ & $158$ $(\sfrac{1}{3})$ \\
STEVE~(GPT-4) & 6 $\pm$ 2 $(\sfrac{3}{3})$ & 10 $\pm$ 1 $(\sfrac{3}{3})$ & \textbf{14} $\pm$ \textbf{3} $\textbf{(\sfrac{3}{3})}$ & \textbf{89} $\pm$ \textbf{9} $\textbf{(\sfrac{3}{3})}$\\
\textbf{STEVE~(Ours)} & \textbf{4} $\pm$ \textbf{1} $\textbf{(\sfrac{3}{3})}$ & \textbf{8} $\pm$ \textbf{1} $\textbf{(\sfrac{3}{3})}$ & $15 \pm 2$ $(\sfrac{3}{3})$ & $106 \pm 12$ $(\sfrac{3}{3})$ \\
\bottomrule
\end{tabular}
}
\label{table:ablation}
\end{table}

To understand the impact of different components on the performance of our system, we conducted ablation studies focusing on the tech tree mastery task in Minecraft. The results, as shown in~\cref{table:ablation}, provide insights into the effectiveness of the vision unit and compare our STEVE model with the STEVE GPT-4 version~(with the same vision unit as ours). Note that the w/o vision unit setup is that the environmental perception encompasses data on blocks within an 8x8 area surrounding the agent, including the front, back, left, and right directions. The following observations are made:

\noindent\paragraph{Vision unit is critical.}
The omission of the vision unit markedly affects the system's performance, especially in more advanced tasks. While it successfully crafts Wooden, Stone, and Iron Tools, it is challenged with Diamond Tools. This outcome underscores the vital importance of visual information in accomplishing complex tasks.

\noindent\paragraph{Comparison with GPT-4.}
As our vision encoder directly encodes into text space, it can be easily replaced with any language model. For instance, the GPT-4 we compared exhibits consistent success across all categories and secures a flawless success rate. Interestingly, the STEVE-13B version excels in simpler tasks such as crafting wooden and stone tools. Moreover, it requires fewer iterations than methods without the vision part, underscoring its superior efficiency.

\subsection{Case Study}

As shown in~\cref{fig:qa}, we perform an extensive case study comparison of GPT-4, LLaMA2-13B, and our method STEVE-13B. Each model maintains the same information and question inputs to compare feedback under different environmental information. Our STEVE overall achieves the best results, surpassing GPT-4 and showing significant improvement compared to the original LLaMA. Especially in parts involving numerical calculations, such as the leftmost image, STEVE accurately tracks food values to restore hunger levels.

\section{Conclusion}

STEVE enhances multi-modal learning by combining visual encoder and LLM-based agents. It has three functions: vision perception, language instruction, and code action, allowing it to understand, predict, and act in virtual environments. We provide a straightforward approach to creating a robust, multi-modal, autonomous, embodied agent using an open-source language model with a small number of parameters. Additionally, we provide a comprehensive dataset STEVE-21K for sustainable community development that can be verified. 

\section*{Acknowledgement}
This work is supported by the Zhejiang Provincial Natural Science Foundation of China (No. LZ24F030005) and the National Natural Science Foundation of China (No. 62106219).

\bibliographystyle{splncs04}
\bibliography{main}
\end{document}